\documentclass[12pt]{article}
\usepackage{graphicx}
\usepackage{amsmath}
\usepackage{algorithm}
\usepackage{algpseudocode}
\usepackage{comment}
\usepackage[numbers]{natbib}
\usepackage{hyperref}
\usepackage{xcolor}
\usepackage{listings}
\usepackage{float}
\usepackage{booktabs}
\usepackage{caption}
\usepackage{subcaption}
\usepackage{tcolorbox}
\usepackage{amssymb}

\title{\textbf{{\LARGE Mind the Gap}}\\[1ex]
\large A Generalized Approach for Cross-Modal Embedding Alignment}

\author{
\begin{minipage}[t]{0.48\textwidth}
    \centering
    \small Arihan Yadav \\
    \small University of Wisconsin-Madison \\
    \texttt{\small apyadav@wisc.edu}
\end{minipage}
\hfill
\begin{minipage}[t]{0.48\textwidth}
    \centering
    \small Alan B. McMillan \\
    \small University of Wisconsin-Madison \\
    \texttt{\small abmcmillan@wisc.edu}
\end{minipage}
}

\begin{document}

\maketitle
\vspace{-1cm}
\begin{abstract}
Retrieval-Augmented Generation (RAG) systems enhance text generation by incorporating external knowledge but often struggle when retrieving context across different text modalities due to semantic gaps. We introduce a generalized projection-based method, inspired by adapter modules in transfer learning, that efficiently bridges these gaps between various text types, such as programming code and pseudocode, or English and French sentences. Our approach emphasizes speed, accuracy, and data efficiency, requiring minimal resources for training and inference. By aligning embeddings from heterogeneous text modalities into a unified space through a lightweight projection network, our model significantly outperforms traditional retrieval methods like the Okapi BM25 algorithm and models like Dense Passage Retrieval (DPR), while approaching the accuracy of Sentence Transformers. Extensive evaluations demonstrate the effectiveness and generalizability of our method across different tasks, highlighting its potential for real-time, resource-constrained applications.
\end{abstract}

\section{Introduction}

Retrieval-Augmented Generation (RAG) systems \citep{lewis2020retrieval} have gained significant attention due to their ability to improve text generation by integrating external knowledge through retrieval mechanisms. These systems enhance the contextual relevance of prompts, addressing the inherent knowledge limitations of generative models. However, RAG systems frequently encounter challenges when bridging the semantic gap between heterogeneous text modalities, such as programming code and pseudocode or between different languages like English and French. This gap complicates the retrieval of contextually relevant information, hindering the system's ability to generate accurate and coherent outputs.

To address this challenge, we propose a novel projection-based method, akin to adapter modules in transfer learning \citep{houlsby2019parameter}, that emphasizes \textbf{speed}, \textbf{data efficiency}, and \textbf{minimal compute resources}. Our approach efficiently aligns embeddings from different text modalities into a unified semantic space through a lightweight projection network, facilitating accurate retrieval across diverse types of text. This alignment significantly reduces training time and data requirements, offering a practical solution for resource-constrained environments. For instance, tasks such as matching English sentences to their French translations or aligning code snippets with corresponding pseudocode can be accomplished with minimal data and computational resources.

Several methods have been proposed to address the challenges of cross-modal retrieval and semantic alignment. Traditional retrieval approaches, such as the Okapi BM25 algorithm \citep{robertson1994some}, are effective for basic text-based searches but struggle to capture nuanced semantic relationships across different modalities or languages. Transformer-based models, such as CodeBERT \citep{feng2020codebert}, have demonstrated potential in representing code and natural language text; however, their performance degrades when faced with substantial differences in structure or vocabulary. Other advancements, including Dense Passage Retrieval (DPR) \citep{karpukhin2020dense} and Sentence Transformers \citep{reimers2019sentence}, aim to enhance retrieval accuracy but often require extensive training data and significant computational resources, limiting their practicality in low-resource or real-time environments.

Our proposed approach introduces a projection-based model designed to overcome these limitations by enabling efficient training on smaller datasets. By projecting embeddings from different modalities into a shared space, similar to how adapters integrate additional tasks into pre-trained models, the model improves the effectiveness of similarity measures used in retrieval tasks. This approach reduces both computational complexity and training time, making it particularly well-suited for applications with constrained data or domain-specific requirements.

\begin{figure}[H]
    \centering
    \includegraphics[width=1\textwidth]{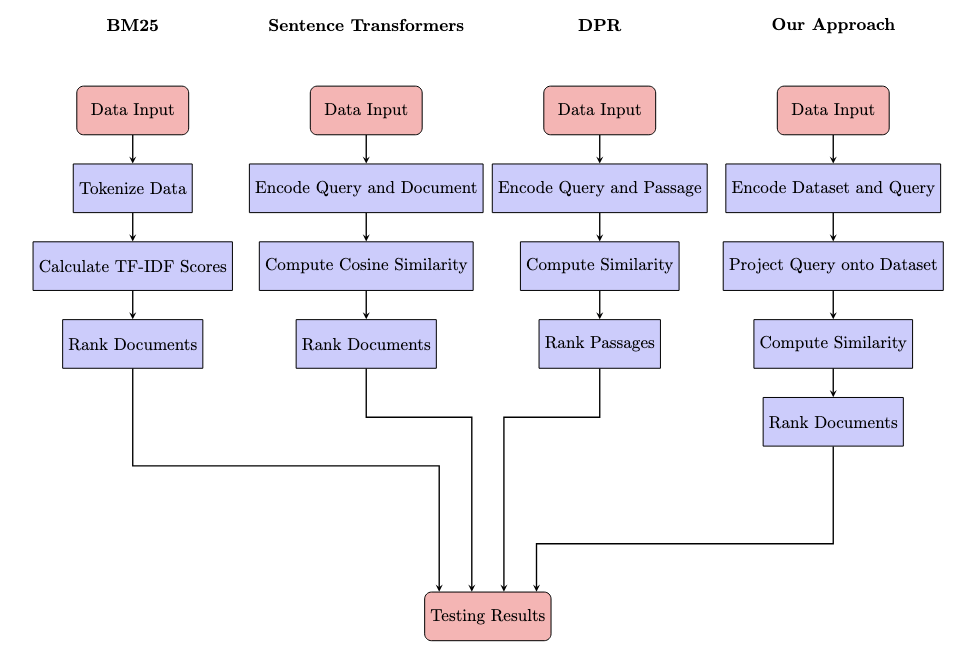}
    \caption{Visual representation of the comparison between the referenced models above and the process followed by our projection-based approach.}
    \label{fig:comparison}
\end{figure}

\section{Methods}

\subsection{Semantic Gap in Heterogeneous Text Modalities}

The semantic gap between distinct text modalities, such as programming code and pseudocode or sentences in different languages, presents a significant challenge when employing direct similarity measures for retrieval tasks. This gap arises due to differences in syntax, structure, and vocabulary between the modalities. As a result, embeddings generated from these different forms of text often fail to align, limiting the effectiveness of traditional similarity metrics, such as cosine similarity, for comparison.

\subsection{Projection-Based Embedding Alignment}

To address this issue, we propose a projection-based model that transforms and aligns embeddings from heterogeneous text types into a unified embedding space. This concept is inspired by adapter modules \citep{houlsby2019parameter}, which are lightweight neural networks inserted into pre-trained models to facilitate transfer learning for new tasks. Similarly, our projection network serves as an adapter between different embedding spaces.

\subsubsection{Neural Network Architecture}

The projection model employs a neural network designed to learn the mapping between embedding spaces of different modalities. The network consists of two Transformer-based encoders and a projection network with one hidden layer.

\paragraph{Encoders}

We utilize pre-trained models from the CodeBERT \citep{feng2020codebert} architecture for encoding both modalities:

\begin{itemize}
    \item \textbf{Modality A Encoder}: Processes inputs from the first modality (e.g., programming code).
    \item \textbf{Modality B Encoder}: Processes inputs from the second modality (e.g., pseudocode).
\end{itemize}

Each encoder generates embeddings of dimension 768, by averaging the last hidden states across the sequence length.

\paragraph{Projection Network}

The projection network acts as an adapter, mapping embeddings from Modality B into the embedding space of Modality A. Its architecture is defined as follows:

\begin{itemize}
    \item \textbf{Input Layer}: A linear transformation from the input dimension 768 to a hidden dimension 2048.
    \item \textbf{First Activation Function}: ReLU activation applied to the output of the input layer.
    \item \textbf{Hidden Layer}: A linear transformation from the hidden dimension 2048 to 2048.
    \item \textbf{Second Activation Function}: ReLU activation applied to the output of the hidden layer.
    \item \textbf{Output Layer}: A linear transformation projecting back to the original embedding dimension 768.
\end{itemize}

\vspace{3mm}

This architecture is implemented as:

\begin{lstlisting}[language=Python]
self.projection = nn.Sequential(
    nn.Linear(768, 2048), # Input layer
    nn.ReLU(),  # First activation function
    nn.Linear(2048, 2048),  # Hidden layer
    nn.ReLU(),  # Second activation function
    nn.Linear(2048, 768)  # Output layer
)
\end{lstlisting}

\vspace{2mm}

This architecture consists of 3 linear layers (input, hidden, and output), with ReLU activations applied after the first and second linear layers. The architecture performs dimensional transformations starting from 768 to 2048, applying activations, and finally projecting back to the 768-dimensional output space.

\begin{figure}[H]
    \centering
    \includegraphics[width=0.7\textwidth]{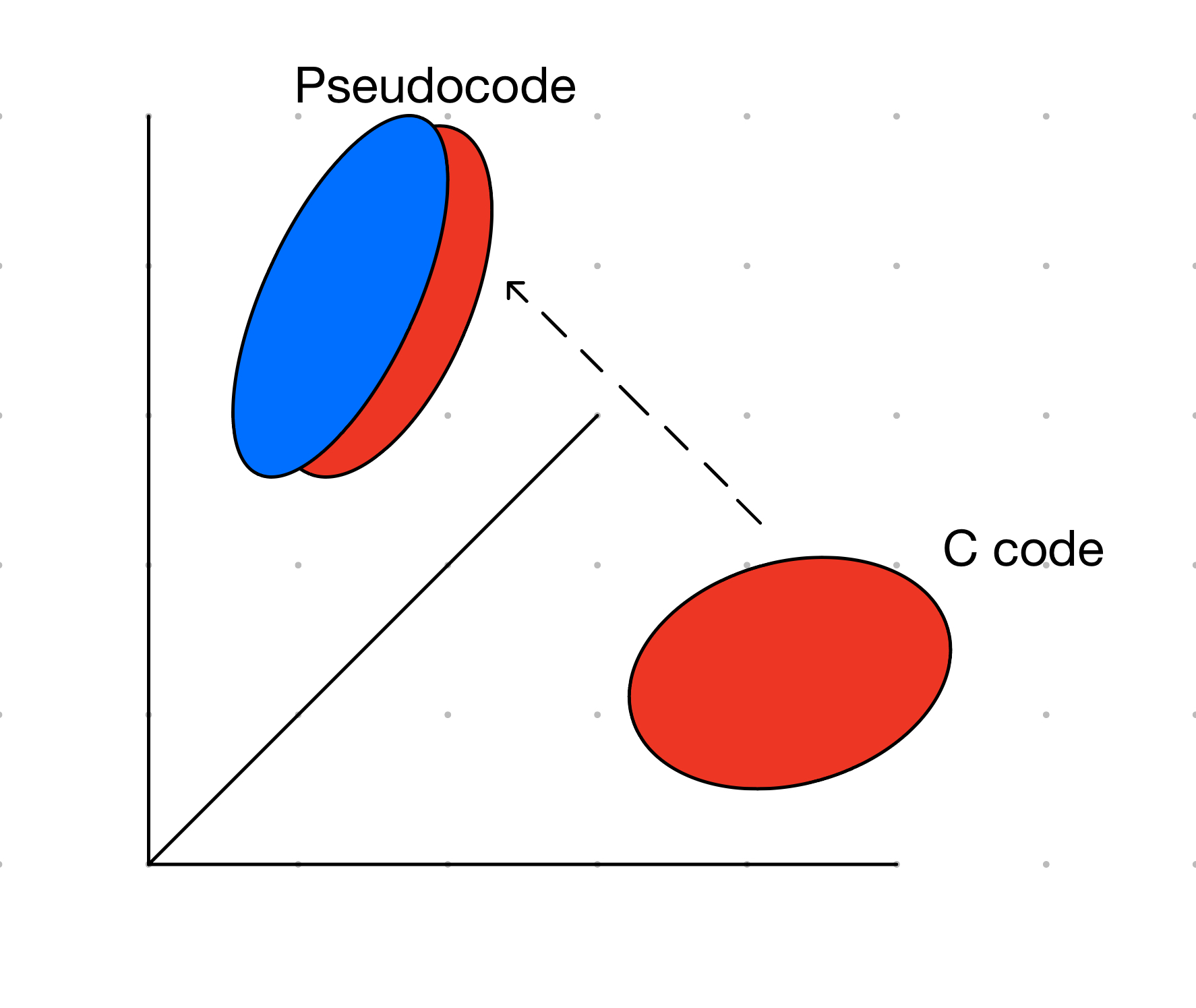}
    \caption{Architecture of the Projection Model aligning embeddings from Modality B (e.g., pseudocode) to Modality A (e.g., programming code) embedding space.}
    \label{fig:embedding-space}
\end{figure}

\subsubsection{Mathematical Formulation}

Let $E_A \in \mathbb{R}^{d}$ be the embedding from Modality A and $E_B \in \mathbb{R}^{d}$ be the embedding from Modality B. The projection network $P$ transforms $E_B$ as follows:

\[
E_B' = P(E_B) = W_3 \cdot \sigma_2\left( W_2 \cdot \sigma_1\left( W_1 \cdot E_B + b_1 \right) + b_2 \right) + b_3
\]

Where:

\begin{itemize}
    \item $W_1 \in \mathbb{R}^{h \times d}$, $W_2 \in \mathbb{R}^{h \times h}$, $W_3 \in \mathbb{R}^{d \times h}$ are weight matrices.
    \item $b_1 \in \mathbb{R}^{h}$, $b_2 \in \mathbb{R}^{h}$, $b_3 \in \mathbb{R}^{d}$ are bias vectors.
    \item $\sigma_1$ and $\sigma_2$ denote the ReLU activation functions.
\end{itemize}

After projection, $E_A$ and $E_B'$ reside in the same embedding space, allowing direct comparison using similarity measures like cosine similarity or Euclidean distance.

\subsubsection{Algorithmic Implementation}

The projection model operates as outlined in Algorithm \ref{alg:projection_model}. This pseudocode abstracts the initialization and forward pass of the model, emphasizing the key computational steps.

\begin{algorithm}[H]
\caption{Projection Model Initialization and Forward Pass}
\label{alg:projection_model}
\begin{algorithmic}[1]
    \State \textbf{Initialize} Modality A Encoder with pre-trained parameters
    \State \textbf{Initialize} Modality B Encoder with pre-trained parameters
    \State \textbf{Initialize} Projection Network $P$ with weights $W_1$, $W_2$, $W_3$ and biases $b_1$, $b_2$, $b_3$
    \State
    \Procedure{Forward}{$\text{Input}_A$, $\text{Input}_B$}
        \State $E_A \gets \text{Modality A Encoder}(\text{Input}_A)$
        \State $E_B \gets \text{Modality B Encoder}(\text{Input}_B)$
        \State $E_B' \gets P(E_B)$ \Comment{Apply projection network}
        \State \textbf{Return} $E_A$, $E_B'$
    \EndProcedure
\end{algorithmic}
\end{algorithm}

\subsubsection{Training Procedure}

During training, the model learns the parameters of the projection network $P$ while keeping the encoders fixed or fine-tuned, depending on the specific experimental setup. The training aims to minimize the distance between the projected embeddings $E_B'$ and the corresponding embeddings $E_A$ for positive pairs, while maximizing the distance for negative pairs.

\subsection{Data Generation and Training}

\subsubsection{Data Preparation}

To train and evaluate our projection model, we created datasets for two different tasks:

\begin{enumerate}
    \item \textbf{Programming Code to Pseudocode}: We used the FormAI dataset \citep{li2023formai} containing C code snippets. Corresponding pseudocode was generated using OpenAI's GPT-3.5 model \citep{openai2023gpt35}, producing both relevant (positive) and irrelevant (negative) pseudocode for each code snippet. This resulted in a balanced dataset of 2,000 positive and negative examples available at \citep{yadav2024github}.
    \item \textbf{English to French Sentences}: We utilized the \newline \texttt{sentence-transformers/wikipedia-en-sentences} dataset from Hugging Face \citep{reimers2019wikipedia}, which contains English sentences extracted from Wikipedia. We then passed these English sentences into OpenAI's GPT-3.5 model \citep{openai2023gpt35} to synthetically generate corresponding French translations, forming positive pairs. Negative examples were created by pairing English sentences with incorrect French translations. The generated dataset of 8,000 positive and negative pairs is available at \citep{yadav2024english}.
\end{enumerate}

\subsubsection{Creating Synthetic Data}

For the code to pseudocode task, we utilized OpenAI's GPT-3.5 model \citep{openai2023gpt35} to generate both relevant and irrelevant pseudocode for each code snippet. This approach ensured that the model could effectively learn to distinguish between accurate and inaccurate pairs. The data generation process involved:

\begin{itemize}
    \item \textbf{Good Pseudocode}: Generating high-level pseudocode that accurately describes the functionality of the code.
    \item \textbf{Bad Pseudocode}: Producing pseudocode that is irrelevant to the given code snippet.
\end{itemize}

Similarly, for the English to French task, we utilized OpenAI's GPT-3.5 model \citep{openai2023gpt35} to generate French translations of English sentences from the \texttt{sentence-transformers/wikipedia-en-sentences} dataset \citep{reimers2019wikipedia}. This allowed us to create positive pairs of English sentences and their corresponding French translations. The generated dataset is available at \citep{yadav2024english}. Negative examples were generated by pairing English sentences with incorrect French translations.

\subsubsection{Training with N-Pairs Loss}

Our model uses a custom \textbf{N-Pairs Loss} function \citep{sohn2016improved} to optimize the alignment of embeddings. The loss function minimizes the distance between positive pairs (e.g., code and corresponding pseudocode) and maximizes the distance between negative pairs. The loss is computed as:

\[
\mathcal{L}_{\text{NPairs}} = \frac{1}{N} \sum_{i=1}^{N} \sum_{j=1, j \neq i}^{N} \max(0, \text{dist}(A_i, P_i) - \text{dist}(A_i, P_j) + \text{margin})
\]

Where:

\begin{itemize}
    \item $A_i$ is the anchor embedding (e.g., code snippet embedding).
    \item $P_i$ is the positive embedding (correct pseudocode embedding).
    \item $P_j$ is the negative embedding (incorrect pseudocode embedding).
    \item $\text{dist}$ is the Euclidean distance metric.
    \item $\text{margin}$ is a hyperparameter set to 1.0.
    \item $N$ is the number of training samples.
\end{itemize}

\begin{figure}[H]
    \centering
    \includegraphics[width=0.7\textwidth]{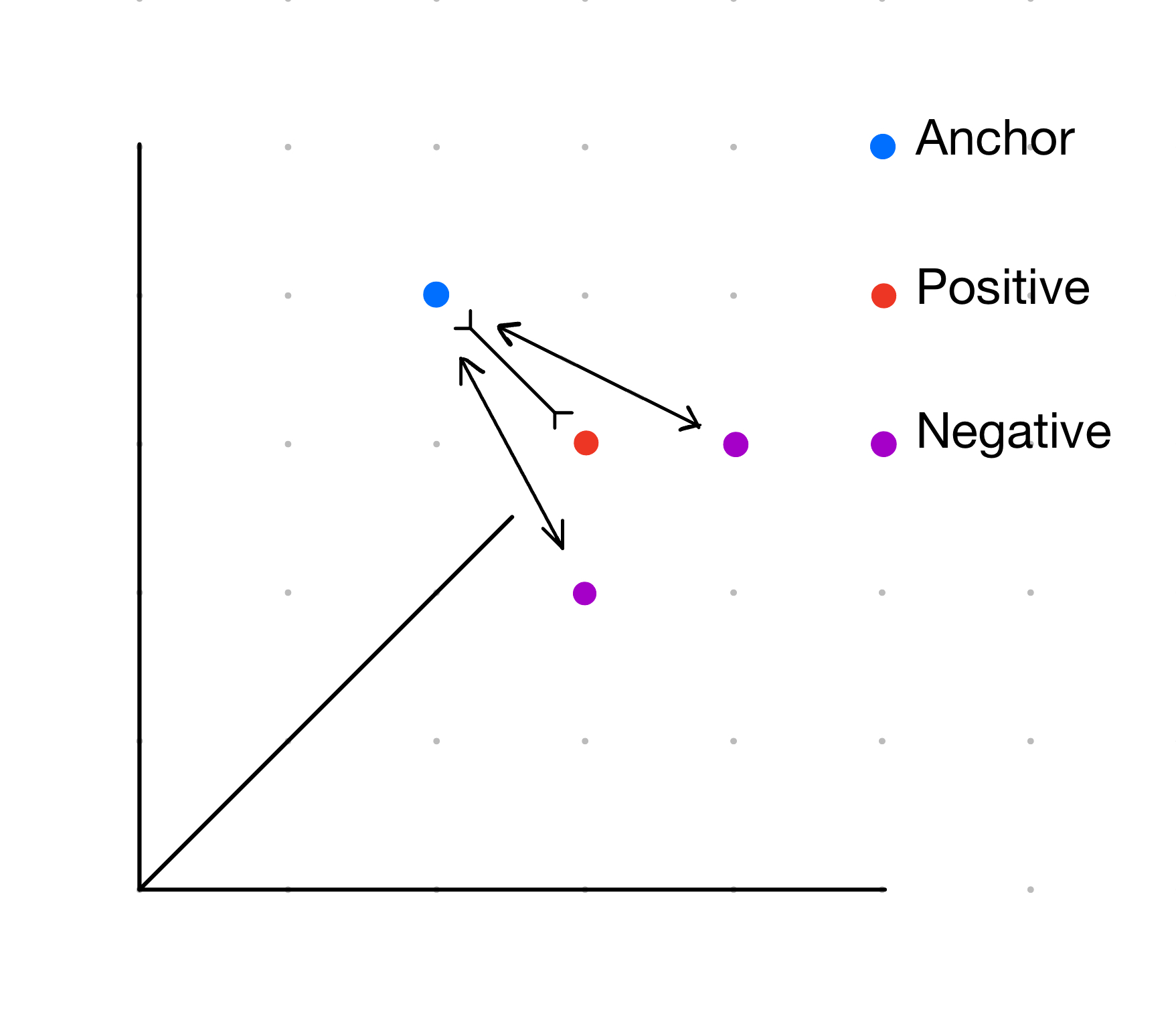}
    \caption{An example representation of positive and negative pairs in the embedding space. The loss is calculated based on the distances between the anchor and the positive point, and the anchor with the negative points.}
    \label{fig:projection_model}
\end{figure}

\begin{algorithm}[H]
\caption{Custom N-Pairs Loss Function}
\label{alg:n_pairs_loss}
\begin{algorithmic}[1]
    \State \textbf{Input:} Anchor embeddings $A$, positive embeddings $P$, labels $\text{labels}$, margin $\text{margin}$
    \State \textbf{Output:} Loss value $\mathcal{L}$
    \State Initialize loss $\mathcal{L} \gets 0.0$
    \State Initialize count $\text{count} \gets 0$
    \For{$i = 1$ to $N$} \Comment{Loop over batch}
        \If{$\text{labels}[i] == 1$} \Comment{Only process positive pairs}
            \State Compute positive pair distance:
            \[
            \text{positive\_distance} \gets \text{dist}(A_i, P_i)
            \]
            \For{$j = 1$ to $N$} \Comment{Compare with other embeddings}
                \If{$i \neq j$}
                    \State Compute negative pair distance:
                    \[
                    \text{negative\_distance} \gets \text{dist}(A_i, P_j)
                    \]
                    \State Update loss:
                    \[
                    \mathcal{L} \gets \mathcal{L} + \max(0, \text{positive\_distance} - \text{negative\_distance} + \text{margin})
                    \]
                    \State Increment count: $\text{count} \gets \text{count} + 1$
                \EndIf
            \EndFor
        \EndIf
    \EndFor
    \If{$\text{count} > 0$}
        \State Return average loss: $\mathcal{L} \gets \frac{\mathcal{L}}{\text{count}}$
    \Else
        \State Return $0.0$
    \EndIf
\end{algorithmic}
\end{algorithm}

Each model was trained for 5 epochs on a single NVIDIA L40 GPU with 40 GB of RAM. Training time 368 seconds for the English to French projection model and 532 seconds for the pseudocode projection model.

\section{Results}

\subsection{Evaluation Metrics}

We compared our model against traditional retrieval methods like \textbf{BM25}, \textbf{Dense Passage Retrieval (DPR)}, and \textbf{Sentence Transformers}. The \textit{all-mpnet-base-v2} version of Sentence Transformers was used as it boasts the highest claimed accuracy of all Sentence Transformers versions. \textbf{All testing was done on a single NVIDIA L40 GPU with 40GB of RAM, with 1000 data pairs in the test set}. The key metrics used in our evaluation are:

\begin{itemize}
    \item \textbf{Accuracy}: The percentage of correct matches retrieved.
    \item \textbf{Precision}: The percentage of retrieved matches that are correct.
    \item \textbf{Recall}: The percentage of correct matches that are retrieved.
    \item \textbf{F1-Score}: The harmonic mean of precision and recall.
\end{itemize} 

\subsection{Performance on English to French Projection}

Table \ref{tab:en-fr-results} presents the evaluation metrics for the English to French projection task. Our projection model achieved the highest accuracy and F1-score among the models tested, demonstrating its effectiveness in aligning embeddings across different languages.

\begin{table}[H]
    \centering
    \caption{Evaluation Metrics for English to French Projection}
    \label{tab:en-fr-results}
    \begin{tabular}{@{}lcccc@{}}
        \toprule
        \textbf{Model} & \textbf{Accuracy} & \textbf{Precision} & \textbf{Recall} & \textbf{F1-Score} \\
        \midrule
        BM25 Model & 0.6800 & 0.6552 & 0.6800 & 0.6591 \\
        DPR Model & 0.6400 & 0.5464 & 0.6400 & 0.5693 \\
        CodeBERT w/o Projection & 0.1640 & 0.0997 & 0.1640 & 0.1102 \\
        Sentence Transformer & 0.9280 & 0.8933 & 0.9280 & 0.9047 \\
        \textbf{Projection Model (Ours)} & 0.9720 & 0.9620 & 0.9720 & 0.9653 \\
        \bottomrule
    \end{tabular}
\end{table}

\subsection{Performance on Code to Pseudocode Projection}

Table \ref{tab:code-pseudo-results} shows the evaluation metrics for the code to pseudocode projection task. Our model significantly outperformed BM25 and DPR models and matched the performance of the Sentence Transformer.

\begin{table}[H]
    \centering
    \caption{Evaluation Metrics for Code to Pseudocode Projection}
    \label{tab:code-pseudo-results}
    \begin{tabular}{@{}lcccc@{}}
        \toprule
        \textbf{Model} & \textbf{Accuracy} & \textbf{Precision} & \textbf{Recall} & \textbf{F1-Score} \\
        \midrule
        BM25 Model & 0.0320 & 0.0320 & 0.0320 & 0.0320 \\
        DPR Model & 0.3840 & 0.3095 & 0.3840 & 0.3252 \\
        CodeBERT w/o Projection & 0.0000 & 0.0000 & 0.0000 & 0.0000 \\
        Sentence Transformer & 0.7840 & 0.7247 & 0.7840 & 0.7432 \\
        \textbf{Projection Model (Ours)} & 0.7840 & 0.7243 & 0.7840 & 0.7427 \\
        \bottomrule
    \end{tabular}
\end{table}

\subsection{Latency and Throughput}

Tables \ref{tab:en-fr-latency} and \ref{tab:code-pseudo-latency} present the average latency per query and the throughput for each model on the English to French and code to pseudocode projection tasks, respectively. Our projection model maintains a low latency, making it suitable for real-time applications.

\begin{table}[H]
    \centering
    \caption{Latency and Throughput for English to French Projection}
    \label{tab:en-fr-latency}
    \begin{tabular}{@{}lcc@{}}
        \toprule
        \textbf{Model} & \textbf{Avg Time (s)} & \textbf{Throughput (Queries/s)} \\
        \midrule
        BM25 Model & 0.0014 & 714.29 \\
        DPR Model & 0.0409 & 24.45 \\
        CodeBERT w/o Projection & 0.0415 & 24.10 \\
        Sentence Transformer & 0.0482 & 20.75 \\
        \textbf{Projection Model (Ours)} & 0.0420 & 23.81 \\
        \bottomrule
    \end{tabular}
\end{table}

\begin{table}[H]
    \centering
    \caption{Latency and Throughput for Code to Pseudocode Projection}
    \label{tab:code-pseudo-latency}
    \begin{tabular}{@{}lcc@{}}
        \toprule
        \textbf{Model} & \textbf{Avg Time (s)} & \textbf{Throughput (Queries/s)} \\
        \midrule
        BM25 Model & 0.0148 & 67.57 \\
        DPR Model & 0.0590 & 16.95 \\
        CodeBERT w/o Projection & 0.0595 & 16.81 \\
        Sentence Transformer & 0.0627 & 15.95 \\
        \textbf{Projection Model (Ours)} & 0.0587 & 17.04 \\
        \bottomrule
    \end{tabular}
\end{table}

\subsection{Balancing Throughput and Accuracy}

To evaluate the trade-off between throughput and accuracy, we compute the harmonic mean between the F1-Score and the throughput (queries per second) for each model. The harmonic mean provides a balanced metric that equally weights the F1-Score and the throughput, highlighting models that perform well on both criteria.

The harmonic mean is calculated as:

\[
\text{Harmonic Mean} = 2 \times \frac{\text{F1-Score} \times \text{Throughput}}{\text{F1-Score} + \text{Throughput}}
\]

\subsubsection{English to French Projection}

Table \ref{tab:en-fr-harmonic} presents the harmonic mean for each model on the English to French projection task.

\begin{table}[H]
    \centering
    \caption{Harmonic Mean for English to French Projection}
    \label{tab:en-fr-harmonic}
    \begin{tabular}{@{}|l|c|@{}}
    \hline
    \textbf{Model} & \textbf{Harmonic Mean} \\ \hline
    BM25 Model & 1.3116 \\ \hline
    DPR Model & 1.1098 \\ \hline
    CodeBERT w/o Projection & 0.2195 \\ \hline
    Sentence Transformer & 1.6822 \\ \hline
    \textbf{Projection Model (Ours)} & \textbf{1.8706} \\ \hline
    \end{tabular}
\end{table}

\subsubsection{Code to Pseudocode Projection}

Table \ref{tab:code-pseudo-harmonic} presents the harmonic mean for each model on the code to pseudocode projection task.

\begin{table}[H]
    \centering
    \caption{Harmonic Mean for Code to Pseudocode Projection}
    \label{tab:code-pseudo-harmonic}
    \begin{tabular}{@{}|l|c|@{}}
    \hline
    \textbf{Model} & \textbf{Harmonic Mean} \\ \hline
    BM25 Model & 0.0640 \\ \hline
    DPR Model & 0.6382 \\ \hline
    CodeBERT w/o Projection & 0.0000 \\ \hline
    Sentence Transformer & 1.4202 \\ \hline
    \textbf{Projection Model (Ours)} & \textbf{1.4233} \\ \hline
    \end{tabular}
\end{table}

\section{Discussion}

The results demonstrate that our projection-based model effectively bridges semantic gaps between heterogeneous text modalities, such as English and French sentences or programming code and pseudocode. In both tasks, the projection model achieved high accuracy and F1-scores, outperforming traditional retrieval methods like BM25 and DPR. Notably, the model's performance closely matches that of the Sentence Transformer while maintaining comparable latency and throughput. \textit{In both tasks, the projection model achieves the highest harmonic mean, indicating a strong balance between accuracy and throughput.}

In the English to French projection task, our model achieved an F1-score of 0.9653 and a harmonic mean of 1.8706, indicating a strong balance between accuracy and throughput. Similarly, in the code to pseudocode projection task, the model attained an F1-score of 0.7427 and a harmonic mean of 1.4233, slightly outperforming the Sentence Transformer in terms of the harmonic mean. The significant improvement over the CodeBERT model without projection highlights the importance of aligning embeddings from different modalities. Directly comparing embeddings from different semantic spaces proved ineffective, as evidenced by the poor performance of the base model without projection. The projection-based approach offers a generalizable solution for cross-modal retrieval tasks. By efficiently aligning embeddings into a unified space through a lightweight projection network, akin to adapter modules \citep{houlsby2019parameter}, the model facilitates accurate retrieval across diverse text types without the need for extensive computational resources or large datasets. This makes it particularly suitable for real-time applications and resource-constrained environments. The ability to effectively align embeddings across different modalities has broad implications in natural language processing and information retrieval. Applications such as cross-lingual retrieval \citep{huang2003cross}, code documentation generation \citep{allamanis2018survey}, and multi-modal data integration could benefit from this approach. Future work could explore extending the projection model to more complex architectures, incorporating advanced adapter mechanisms \citep{pfeiffer2020adapterfusion}, or applying it to other modalities such as audio or visual data. Additionally, investigating the integration of this approach with larger pre-trained models or in combination with other parameter-efficient transfer learning techniques could further enhance performance.

This study has several limitations. First, the approach was evaluated on only two RAG-relevant tasks: cross-language retrieval and pseudocode retrieval. Expanding the evaluation to additional application areas, such as medical literature retrieval, legal document retrieval, or multimedia content retrieval, could reveal further limitations and potential competitive advantages. For instance, medical literature retrieval might highlight challenges in handling domain-specific terminology, while legal document retrieval could test the system’s ability to manage complex legal language and context. Second, the study focused solely on retrieval performance, without integrating a full RAG pipeline. It is important to note that the efficacy of RAG systems is heavily reliant on accurate and timely retrieval. Future work should incorporate the retrieval component into a complete RAG pipeline to assess end-to-end performance, including the generation of responses based on retrieved information. This would help identify any bottlenecks or inefficiencies that might arise when retrieval is combined with other stages of the RAG process. Lastly, the approach necessitates training an additional model with both positive and negative examples. Although the training times were minimal and the data was easily synthesized using the widely available and cost-effective GPT-3.5 model, careful consideration is required to determine the optimal contexts for applying the projection model approach.

\section{Conclusion}

This paper presents a generalized projection-based method for bridging semantic gaps between heterogeneous text modalities. Inspired by adapter modules, we project embeddings from different modalities into a unified semantic space, significantly improving retrieval accuracy while maintaining low latency. Extensive evaluations on tasks involving programming code to pseudocode and English to French sentence alignment demonstrate the effectiveness and generalizability of the approach. The projection model outperforms traditional retrieval methods and approaches the performance of more resource-intensive models like Sentence Transformers. The method's efficiency in training and inference, combined with its minimal data requirements, highlights its potential for deployment in real-time, resource-constrained applications, and lays the groundwork for future research in cross-modal retrieval and embedding alignment.

\newpage

\bibliographystyle{plainnat}
\bibliography{bibliography}

\begin{thebibliography}{15}
\providecommand{\natexlab}[1]{#1}
\providecommand{\url}[1]{\texttt{#1}}
\expandafter\ifx\csname urlstyle\endcsname\relax
  \providecommand{\doi}[1]{doi: #1}\else
  \providecommand{\doi}{doi: \begingroup \urlstyle{rm}\Url}\fi

\bibitem[Allamanis et~al.(2018)Allamanis, Barr, Devanbu, and Sutton]{allamanis2018survey}
Miltiadis Allamanis, Earl~T. Barr, Premkumar Devanbu, and Charles Sutton.
\newblock A survey of machine learning for big code and naturalness.
\newblock \emph{ACM Computing Surveys}, 2018.

\bibitem[Feng et~al.(2020)Feng, Guo, Tang, et~al.]{feng2020codebert}
Zhangyin Feng, Daya Guo, Duyu Tang, et~al.
\newblock Codebert: A pre-trained model for programming and natural languages.
\newblock In \emph{Proceedings of the 2020 Conference on Empirical Methods in Natural Language Processing (EMNLP)}, 2020.

\bibitem[Houlsby et~al.(2019)Houlsby, Giurgiu, Jastrzebski, et~al.]{houlsby2019parameter}
Neil Houlsby, Ana Giurgiu, Stanislaw Jastrzebski, et~al.
\newblock Parameter-efficient transfer learning for nlp.
\newblock In \emph{Proceedings of the 36th International Conference on Machine Learning}, 2019.

\bibitem[Huang and Gao(2003)]{huang2003cross}
Xuedong Huang and Jianfeng Gao.
\newblock Cross-language information retrieval using semantic representation.
\newblock In \emph{Proceedings of the IEEE International Conference on Natural Language Processing and Knowledge Engineering}, 2003.

\bibitem[Karpukhin et~al.(2020)Karpukhin, Oguz, Min, et~al.]{karpukhin2020dense}
Vladimir Karpukhin, Barlas Oguz, Sewon Min, et~al.
\newblock Dense passage retrieval for open-domain question answering.
\newblock In \emph{Proceedings of the 2020 Conference on Empirical Methods in Natural Language Processing (EMNLP)}, 2020.

\bibitem[Lewis et~al.(2020)Lewis, Perez, Piktus, et~al.]{lewis2020retrieval}
Patrick Lewis, Ethan Perez, Aleksandra Piktus, et~al.
\newblock Retrieval-augmented generation for knowledge-intensive nlp tasks.
\newblock In \emph{Advances in Neural Information Processing Systems}, 2020.

\bibitem[Li et~al.(2023)Li, Zhang, and Wang]{li2023formai}
Han Li, Yu~Zhang, and Ji~Wang.
\newblock {FormAI}: A dataset for machine learning in formal methods.
\newblock In \emph{Proceedings of the 31st ACM Joint Meeting on European Software Engineering Conference and Symposium on the Foundations of Software Engineering (ESEC/FSE 2023)}, 2023.

\bibitem[OpenAI(2023)]{openai2023gpt35}
OpenAI.
\newblock {GPT}-3.5: Language models are few-shot learners, 2023.
\newblock OpenAI API.

\bibitem[Pfeiffer et~al.(2020)Pfeiffer, Kamath, Rücklé, Cho, and Gurevych]{pfeiffer2020adapterfusion}
Jonas Pfeiffer, Aishwarya Kamath, Andreas Rücklé, Kyunghyun Cho, and Iryna Gurevych.
\newblock Adapterfusion: Non-destructive task composition for transfer learning.
\newblock In \emph{Proceedings of the 58th Annual Meeting of the Association for Computational Linguistics}, 2020.

\bibitem[Reimers and Gurevych(2019{\natexlab{a}})]{reimers2019sentence}
Nils Reimers and Iryna Gurevych.
\newblock Sentence-bert: Sentence embeddings using siamese bert-networks.
\newblock In \emph{Proceedings of the 2019 Conference on Empirical Methods in Natural Language Processing}, 2019{\natexlab{a}}.

\bibitem[Reimers and Gurevych(2019{\natexlab{b}})]{reimers2019wikipedia}
Nils Reimers and Iryna Gurevych.
\newblock {sentence-transformers/wikipedia-en-sentences} dataset, 2019{\natexlab{b}}.
\newblock Available at \url{https://huggingface.co/datasets/sentence-transformers/wikipedia-en-sentences}.

\bibitem[Robertson and Walker(1994)]{robertson1994some}
Stephen Robertson and Steve Walker.
\newblock Some simple effective approximations to the 2-poisson model for probabilistic weighted retrieval.
\newblock In \emph{Proceedings of the 17th Annual International ACM SIGIR Conference on Research and Development in Information Retrieval}, 1994.

\bibitem[Sohn(2016)]{sohn2016improved}
Kihyuk Sohn.
\newblock Improved deep metric learning with multi-class n-pair loss objective.
\newblock In \emph{Advances in Neural Information Processing Systems}, 2016.

\bibitem[Yadav and McMillan(2024{\natexlab{a}})]{yadav2024english}
Arihan Yadav and Alan~B. McMillan.
\newblock English-french-translations-train-large dataset, 2024{\natexlab{a}}.
\newblock Available at \url{https://huggingface.co/datasets/aircrypto/English-French-Translations-Train-Large}.

\bibitem[Yadav and McMillan(2024{\natexlab{b}})]{yadav2024github}
Arihan Yadav and Alan~B. McMillan.
\newblock {GitHub-C-Code-Pseudocode-C-Py-New-v3} dataset, 2024{\natexlab{b}}.
\newblock Available at \url{https://huggingface.co/datasets/aircrypto/GitHub-C-Code-Pseudocode-C-Py-New-v3}.

\end{thebibliography}

\end{document}